\documentclass[final]{cvpr}

\usepackage{times}
\usepackage{epsfig}
\usepackage{graphicx}
\usepackage{amsmath}
\usepackage{amssymb}
\usepackage{booktabs}
\usepackage{multirow}
\usepackage{color}
\usepackage{lipsum}
\usepackage{xcolor}
\usepackage{commath}
\usepackage{colortbl}
\usepackage{wrapfig}
\usepackage{slashbox}
\usepackage[ruled]{algorithm2e}
\usepackage{gensymb}
\usepackage{enumerate}

\usepackage[position=top]{subfig}
\usepackage{xspace}
\usepackage{booktabs}
\usepackage{tabulary}
\usepackage{floatrow}
\usepackage{wrapfig}
\usepackage{appendix}
\usepackage[demo,abs]{overpic}

\definecolor{citecolor}{RGB}{34,139,34}
\definecolor{grayDark}{gray}{0.95}
\definecolor{grayLight}{gray}{0.98}
\definecolor{darkgreen}{rgb}{0.10, 0.55, 0.10}
\newcommand\green[1]{\textcolor{darkgreen}{#1}}

\newcommand*{\plottitle}{\bf\fontfamily{phv}\fontsize{6}{9}\selectfont}

\usepackage[pagebackref=true,breaklinks=true,letterpaper=true,colorlinks,
  citecolor=citecolor,bookmarks=false]{hyperref}

\usepackage{colortbl}
\usepackage{bm}
\usepackage{algpseudocode}

\newcommand{\et}{\emph{et al.}}

\newcommand{\myparagraph}[1]{{ \noindent \bf #1}}

\def\cvprPaperID{902} % *** Enter the CVPR Paper ID here
\def\confYear{CVPR 2021}

\begin{document}

%%%%%%%%% TITLE
\title{PoseAug: A Differentiable Pose Augmentation Framework for 3D Human Pose Estimation}

\author{
Kehong Gong    \footnotemark[1] $^{\ \dagger}$ \qquad
Jianfeng Zhang \thanks{Equal contribution; order determined by coin toss. $^{\dagger}$Work done during an internship at Huawei international Pte Ltd.}   \qquad
Jiashi Feng \\
National University of Singapore\\
\small \texttt{gongkehong@u.nus.edu} \quad \texttt{zhangjianfeng@u.nus.edu} \quad \texttt{elefjia@nus.edu.sg}
}

\maketitle

%%%%%%%%% ABSTRACT
\begin{abstract}
Existing 3D human pose estimators suffer poor generalization performance to new datasets, largely due to the limited diversity of 2D-3D pose pairs in the training data.
To address this problem, we present PoseAug, a new auto-augmentation framework that learns to augment the available training poses towards a greater diversity and thus improve  generalization of the trained 2D-to-3D pose estimator. 
Specifically, PoseAug introduces a novel pose augmentor that learns to adjust various geometry factors (\eg, posture, body size, view point and position) of a pose through differentiable operations.
With such differentiable capacity, the augmentor can be jointly optimized with the 3D pose estimator and take the estimation error as  feedback to generate more diverse and harder poses in an online manner.  
Moreover, PoseAug introduces a novel part-aware Kinematic Chain Space for evaluating local joint-angle plausibility and develops a discriminative module accordingly to ensure the plausibility of the augmented poses. 
These elaborate designs enable PoseAug to generate more diverse yet plausible poses than existing offline augmentation methods, and thus yield better generalization of the pose estimator.  
PoseAug is generic and easy to be applied to various 3D pose estimators. 
Extensive experiments demonstrate that PoseAug  brings clear improvements on both intra-scenario and cross-scenario  datasets. 
Notably, it achieves 88.6\% 3D PCK on MPI-INF-3DHP under cross-dataset evaluation setup, improving upon the previous best data augmentation based method~\cite{Li_2020_CVPR} by 9.1\%. Code can be found at: \url{https://github.com/jfzhang95/PoseAug}.

\end{abstract}

\section{Introduction}

\begin{figure}[!t]
  \centering
  \subfloat[{\plottitle ~~~~Source dataset: H36M}]{\includegraphics[width=0.5\linewidth]{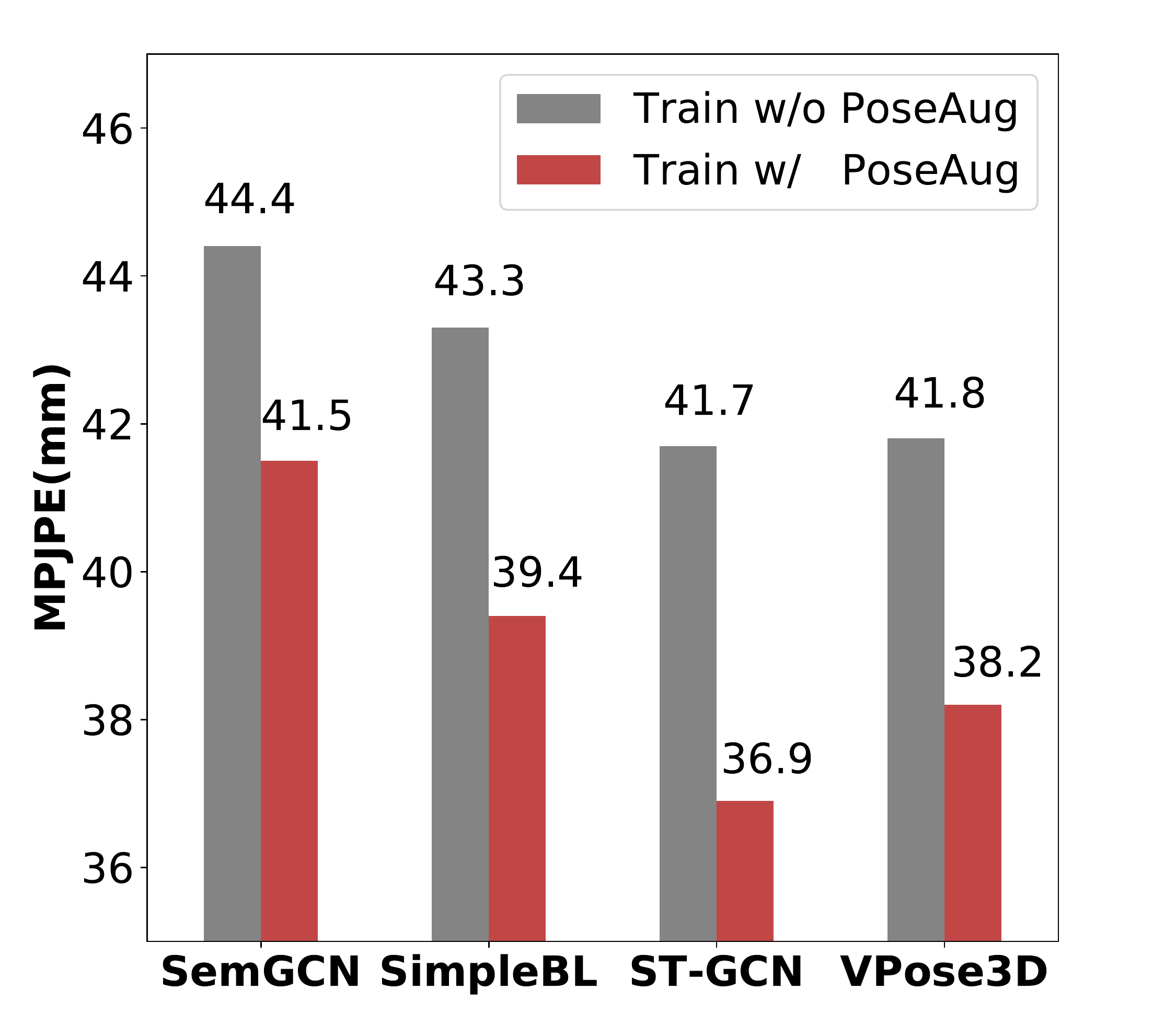}}
  \subfloat[{\plottitle ~~~~Cross dataset: 3DHP}]{\includegraphics[width=0.5\linewidth]{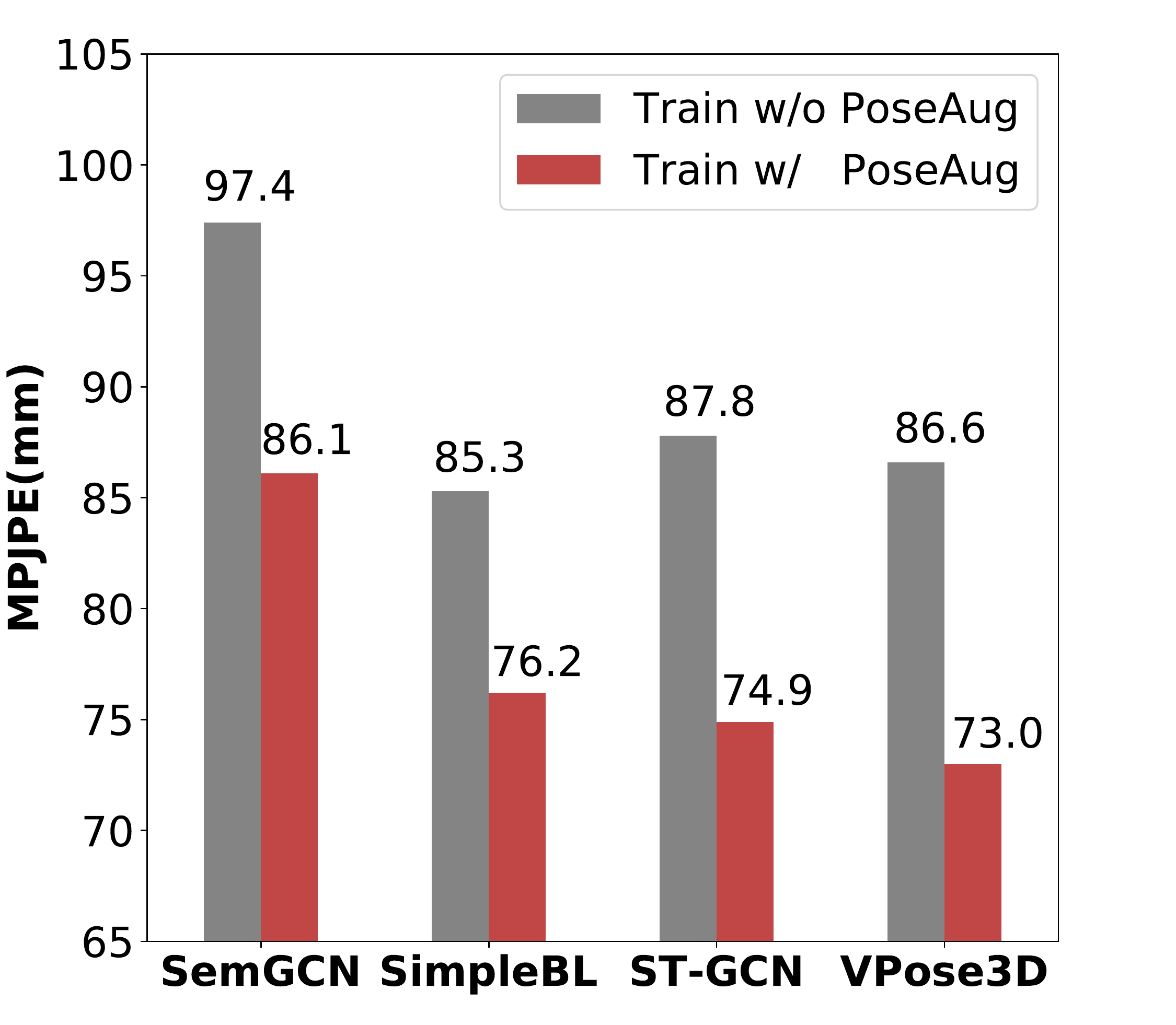}}
  \vspace{-0.5em}
  \caption{Estimation error (in MPJPE) on H36M (intra-dataset evaluation) and 3DHP (cross-dataset evaluation) of four well established models~\cite{zhao2019semantic,martinez2017simple,pavllo2019videopose3d,cai2019exploiting} trained with and  without PoseAug.  PoseAug significantly improves  their performance for both the intra- and cross-dataset settings.}
\label{fig:fig1}
\end{figure}

3D human pose estimation aims to estimate 3D body joints in images or videos. 
It is a fundamental task with broad applications in action recognition~\cite{yan2018spatial,si2019attention}, human-robot interaction~\cite{errity2016human}, human tracking~\cite{mehta2017vnect}, \etc. 
This task is typically solved using learning-based methods~\cite{martinez2017simple,zhao2019semantic,cai2019exploiting,nie2019spm} with ground truth annotations that are collected in the laboratorial environments~\cite{ionescu2014human3}.
Despite their success in indoor scenarios, these methods are hardly generalizable to cross-scenario datasets (\eg, an in-the-wild dataset). 
We argue that their poor generalization is mainly due to the limited diversity of training data, such as limited variations in human posture, body size, camera view point and position.

Recent works explore data augmentation to improve the training data diversity and  enhance the generalization of their trained models.
They either generate data through image composition~\cite{rogez2016mocap,mehta2017vnect,singleshotmultiperson2018} and synthesis~\cite{chen2016synthesizing,varol2017learning}, or directly generate 2D-3D pose pairs from the available training data by applying pre-defined transformations~\cite{Li_2020_CVPR}.
However, all of these works regard data augmentation and model training as two separate phases, and conduct data augmentation in an offline manner without interaction with model training. 
Consequently, they tend to generate ineffective augmented data that are too easy for model training, leading to marginal boost to the model generalization.
Moreover, these methods heavily rely on pre-defined rules such as joint angle limitations~\cite{akhter2015joint_angle_limit} and kinematics constraints~\cite{rogez2016mocap} for data augmentation, which limit the diversity of the generated data and make the resulting model hardly generalize to more challenging in-the-wild scenes.

To improve the diversity of augmented data, we propose \textit{PoseAug}, a novel auto-augmentation framework for 3D human pose estimation.
Instead of conducting data augmentation and network training separately, PoseAug jointly optimizes the augmentation process with network training end-to-end in an online manner.
Our main insight is that the feedback from the training process can be used as effective guidance signals to adapt and improve the data augmentation.
Specifically, PoseAug exploits a differentiable augmentation module (the `augmentor') {implemented by a neural network} to directly augment 2D-3D pose pairs in the training data.
Considering the potential domain shift with respective to geometry in pose pairs (\eg, postures, view points)~\cite{rhodin2018unsupervised,Li_2020_CVPR,zhang2020inference}, the augmentor learns to perform three types of augmentation operations to respectively control 1) the skeleton joint angle, 2) the body size, and 3) the view point and human position.
In this way, the augmentor is able to produce augmented poses with more diverse geometric features and thus relieves the diversity limitation issue.
With its differentiable capacity, the augmentor can be optimized together with the pose estimator end-to-end via an error feedback strategy.
Concretely, by taking increasing training loss of the estimator as the learning target, the augmentor can learn to enrich the input pose pairs via enlarging data variations and difficulties; in turn, through combating such increasing difficulties, the pose estimator can become increasingly more powerful during the training process.

\begin{figure}[!t]
\centering
\includegraphics[width=0.95\linewidth]{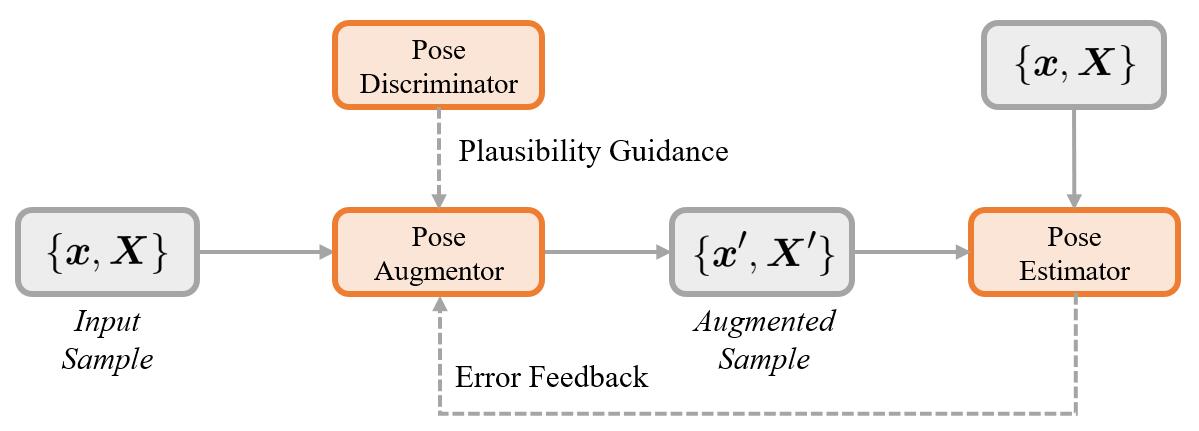}
\caption{\textbf{Overview of our PoseAug framework}. The augmentor, estimator and discriminator are jointly trained  end-to-end with an error-feedback training strategy. 
As such, the augmentor learns to augment data with  guidance from the estimator and   discriminator.}
\label{fig:overview}
\end{figure}

To ensure the plausibility of the augmented poses, we use a pose discriminator module to guide the augmentation, to avoid generating implausible joint angles~\cite{akhter2015joint_angle_limit}, unreasonable positions or view points that may hamper model training. In particular, the module consists of a 3D pose discriminator for enhancing the joint angle plausibility and a 2D pose discriminator for guiding the body size, view point and position plausibility. The 3D pose discriminator adopts the Kinematic Chain Space (KCS)~\cite{wandt2019repnet} representation and extends it into a \textit{part-aware} KCS for local-wise supervision. More concretely, it splits skeleton joints into several parts and focuses on joint angles in each part separately instead of the whole body pose, which yields greater flexibility of the augmented poses. 
By jointly training the pose augmentor, estimator and discriminator in an end-to-end manner (Fig.~\ref{fig:overview}), PoseAug can largely improve the training data diversity, and thus boost model performance on both source and more challenging cross-scenario datasets.

Our PoseAug framework is flexible regarding the choice of the 3D human pose estimator.
This is demonstrated by the clear improvements made with PoseAug on four representative 3D pose estimation models~\cite{zhao2019semantic,martinez2017simple,pavllo2019videopose3d,cai2019exploiting} over both source (H36M)~\cite{ionescu2014human3} and cross-scenario (3DHP)~\cite{mehta2017vnect} datasets (Fig.~\ref{fig:fig1}).
Remarkably, it brings more than 13.1\% average improvement w.r.t. MPJPE for all models  on 3DHP.
Moreover, it achieves 88.6\% 3D PCK on 3DHP under cross-dataset evaluation setup, improving upon the previous best data augmentation based method~\cite{Li_2020_CVPR} by 9.1\%.

Our contributions are three-fold. 
1) To the best of our knowledge, we are the first to investigate differentiable data augmentation on 3D human pose estimation. 
2) We propose a differentiable pose augmentor, together with the error feedback design, which generates diverse and realistic 2D-3D pose pairs for training the 3D pose estimator, and largely enhances the model's generalization ability.
3) We propose a new \textit{part-aware} 3D discriminator, which enlarges the feasible region of augmented poses via local-wise supervision, ensuring both data plausibility and diversity.

\section{Related Work}

\myparagraph{3D human pose estimation}
Recent   progress of 3D human pose estimation is largely driven by the deployment of various deep neural network   models~\cite{tekin2016direct,martinez2017simple,fang2018learning,zhao2019semantic,nibali20193d,cai2019exploiting,sharma2019monocular,zhou2019hemlets}. However, they all   highly rely on well-annotated data for fully-supervised  model training and hardly generalize to the new scenarios that present unseen   patterns in the training dataset, such as new camera views and subject poses. 
Thus some recent works explore to leverage external information to improve their generalization ability. For example, some methods~\cite{zhou2017towards,yang20183d,dabral2018learning,wandt2019repnet,inthewild3d_2019,wang2019generalizing,chen2019weakly,pavllo2019videopose3d,kolotouros2019spin} utilize 2D pose  data collected  in the wild  for model training, \eg, through   exploring  kinematics priors for regularization or post-processing~\cite{zhou2017towards,dabral2018learning,pavllo2019videopose3d}, and adversarial training~\cite{yang20183d,wandt2019repnet}.
More recently, geometry-based self-supervised learning~\cite{rhodin2018unsupervised,drover2018can,chen2019unsupervised,kocabas2019epipolar,pirinen2019domes,ligeometry,rhodin2019neural} has been used 
to train models with   unlabeled {data}. 
Though effective, applying these methods is largely constrained by the availability of suitable external datasets. 
Instead of focusing on complex network architectures and learning schemes, we explore a learnable pose augmentation framework to enrich the 3D pose data at hand directly. Specifically, the proposed framework can generate 2D-3D pose pairs with both diversity and plausibility for training pose estimation models.
In addition, our framework is generic and can adapt to those methods to further improve their performance.

\myparagraph{Data augmentation on 3D human poses}
Data augmentation is widely used to alleviate the bottleneck of training data diversity and improve model generalization ability.
Some works augment data by stitching image patches~\cite{rogez2016mocap,mehta2017vnect,zhang2021bmp}, and some generate new data with graphics engines~\cite{chen2016synthesizing,varol2017learning}. 
More recently,
Li~\et,~\cite{Li_2020_CVPR} directly augment 2D-3D pose pairs through randomly applying partial skeleton recombination and joint angle perturbation on source datasets.
To ensure data plausibility, several constraints are imposed, including joint angle limitation~\cite{akhter2015joint_angle_limit} and fixed augmentation range on view point and human position.
Despite the good results on source data, these pre-defined rules limit the data diversity expansion and harm the model applicability to more challenging in-the-wild scenarios.
Unlike all these methods, we make the first attempt to explore learnable data augmentation on 3D human pose estimation, which is shown effective for improving model generalization ability.

\section{Method}

\subsection{Problem Definition}

Let $\boldsymbol{x} \in \mathbb{R}^{2\times J}$ denote 2D spatial coordinates of $J$ keypoints of the human in the image, and $\boldsymbol{X} \in \mathbb{R}^{3\times J}$ denote the corresponding 3D joint position in the camera coordinate system.
We aim to obtain a 3D pose estimator $\mathcal{P}: \boldsymbol{x} \mapsto \boldsymbol{X}$ to recover the 3D pose information from the input 2D  pose. Conventionally, the estimator $\mathcal{P}$, with parameters $\theta$, is trained on a well-annotated source dataset (\eg, well-controlled indoor environment~\cite{ionescu2014human3}) by solving the following optimization problem:
\begin{align}
\min_{\theta} \mathcal{L}_{\mathcal{P}}(\mathcal{P}_\theta, \mathcal{X}) =  \mathcal{L}_{\mathcal{P}}(\mathcal{P}_\theta(\boldsymbol{x}), \boldsymbol{X}), 
\end{align}
where $\mathcal{X} = \{\boldsymbol{x}, \boldsymbol{X}\}$ denotes paired 2D-3D poses from the source training dataset, and the loss function $\mathcal{L}_{\mathcal{P}}$ is typically defined as mean square errors (MSE) between predicted and ground truth 3D poses.
However, it is often observed that the pose estimator $\mathcal{P}$ trained on such an indoor dataset can hardly generalize to a  new dataset (\eg, in-the-wild scenario) which features more diverse poses, body sizes, view points or human positions~\cite{inthewild3d_2019,zhang2020inference,wang2020predicting}.

To improve generalization ability of the model, we propose to design a pose augmentor $\mathcal{A}: \mathcal{X} \mapsto \mathcal{X}' $, to augment the training pose pair $\mathcal{X}$ into a more diverse one $\mathcal{X}'=\{\boldsymbol{x}', \boldsymbol{X}'\}$ for  training the model $\mathcal{P}$:
\begin{equation}
    \min_{\theta}  \mathcal{L}_{\mathcal{P}}(\mathcal{P}_\theta,   \mathcal{A}(\mathcal{X})).
\end{equation}
There are several strategies to construct the augmentor in an offline manner, \eg, random~\cite{chen2016synthesizing,mehta2017vnect,varol2017learning} or evolution-based augmentations~\cite{Li_2020_CVPR}.
Differently, we propose to implement the augmentor $\mathcal{A}$ via a neural network with parameters $\theta_A$ and train it jointly with the estimator in an online manner, such that the pose estimator loss can be fully exploited as a surrogate for the augmentation diversity and effectively guide the augmentor learning. In particular, the augmentor is trained to generate harder augmented samples that could increase the training loss of the current pose estimator: 
\begin{equation}
\label{eq:aug_train}
    \min_{\theta} \max_{\theta_A} \mathcal{L}_{\mathcal{P}}(\mathcal{P}_\theta, \mathcal{A}_{\theta_A}(\mathcal{X})).
\end{equation}

\subsection{PoseAug Formulation}

Our proposed framework aims to generate diverse training data, with proper difficulties for the pose estimator, to improve model generalization performance.
Two challenges thus need to be tackled: how to make the augmented data diverse and beneficial for model training; and how to make them natural and realistic. 
To address them, we propose two novel ideas in training the augmentor.

\myparagraph{Error feedback learning for online pose augmentation}
Instead of performing random pose augmentation in an offline manner \cite{rogez2016mocap,chen2016synthesizing,Li_2020_CVPR}, the proposed pose augmentator $\mathcal{A}$ deploys  a differentiable design which enables online joint-training with the pose estimator $\mathcal{P}$. 
Using the training error from the pose estimator $\mathcal{P}$ as feedback (see Eqn.~\eqref{eq:aug_train}), the pose augmentor $\mathcal{A}$ learns to generate poses that are most suitable for the current pose estimator\textemdash the augmented poses present proper   difficulties and diversity due to online augmentation, thus maximally benefiting generalization of the trained 3D pose estimation model.

\myparagraph{Discriminative learning for plausible pose augmentation}
Purely pursuing error-maximized augmentations may result in implausible training poses that violate the bio-mechanical structure of human body and may hurt model performance.
Previous augmentation methods~\cite{rogez2016mocap,chen2016synthesizing, Li_2020_CVPR} mostly rely on pre-defined rules for ensuring plausibility  (\eg, joint angle constraint~\cite{akhter2015joint_angle_limit}), which however would severely limit the diversity of generated poses.  
For example, some harder yet plausible  poses  may fail to pass  their rule-based  plausibility check~\cite{Li_2020_CVPR} and will not be adopted for model training.  
To address this issue, we deploy a pose discriminator module over the local relation of body joints~\cite{wandt2019repnet} to assist training the augmentor, thus ensuring the plausibility of augmented poses without sacrificing the diversity.

\subsection{Architecture} 
\label{sec:network}
Fig.~\ref{fig:overview} summarizes our PoseAug architecture design. It includes 
1) a pose augmentor that augments the input pose pair $\{\boldsymbol{x}, \boldsymbol{X}\}$ {to an augmented one} $\{\boldsymbol{x}', \boldsymbol{X}'\}$ for pose estimator $\mathcal{P}$ training; 
2) a pose discriminator module with two discriminators in 3D and 2D spaces, to ensure the plausibility of the augmented data; and 
3) a 3D pose estimator, that provides pose estimation error feedback. 

\myparagraph{Augmentor}
Given a 3D pose
$\boldsymbol{X}\in \mathbb{R}^{3\times J}$, the augmentor first obtains its bone vector $\boldsymbol{B} \in \mathbb{R}^{3 \times (J-1) }$ via a hierarchical transformation\footnote{The hierarchical transformation converts the $J$ joints of $\boldsymbol{X}$ into $J-1$ column vectors of $\boldsymbol{B}$, each of which represents a line segment connecting two adjacent joints.}  $\boldsymbol{B}=\mathcal{H}(\boldsymbol{X})$~\cite{wandt2019repnet,Li_2020_CVPR},   which can be further decomposed into a bone direction vector $\hat{\boldsymbol{B}}$ (representing the joint angle)  and a bone length vector $\|\boldsymbol{B}\|$ (representing the body size).

\begin{figure}[!t]
\centering
\includegraphics[width=0.8\linewidth]{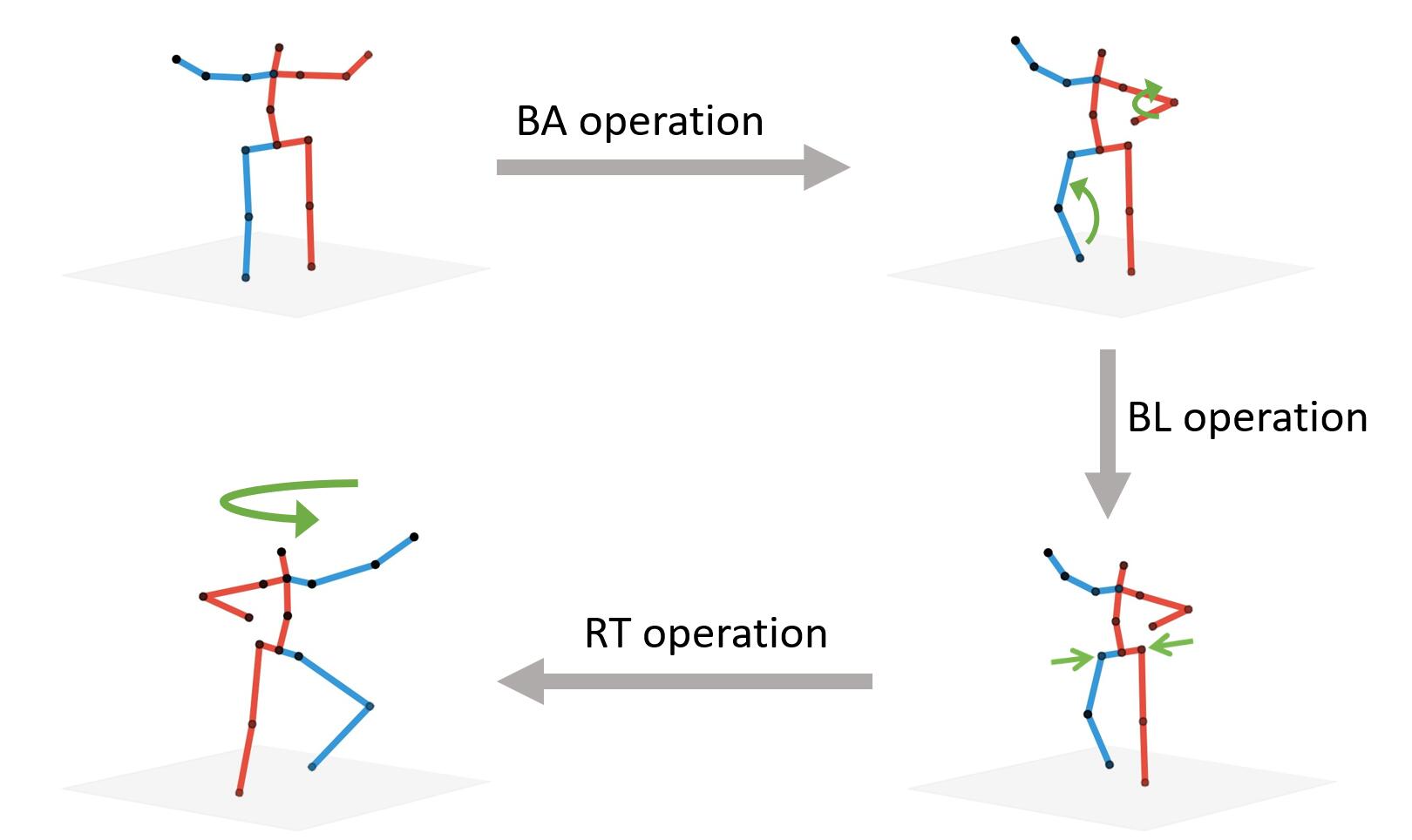}
\caption{\textbf{Augmentation operations with PoseAug.} 
A source 3D pose is augmented by modifying its posture (via BA operation), body size (via BL operation) and view point and position (via RT operation).}
\label{fig:augmentor}
\end{figure}

Then the augmentor applies  multi-layer perceptron (MLP) 
for feature extraction from the input 3D pose $\boldsymbol{X}$.
Additionally, a noise vector based on Gaussian distribution  
is concatenated with $\boldsymbol{X}$ in the feature extraction process to incur sufficient randomness for enhancing the feature diversity. 
The extracted features are then used for regressing three operation parameters ($\boldsymbol{\gamma_{ba}}$, $\boldsymbol{\gamma_{bl}}$ and ($\boldsymbol{R}$, $\boldsymbol{t}$)) to change the joint angles, body size, as well as view point and position as illustrated in Fig.~\ref{fig:augmentor}. Among these parameters,
\begin{enumerate} [1)]
\item  $\boldsymbol{\gamma_{ba}} \in \mathbb{R}^{3\times (J-1)}$ is the bone angle residual vector that is used for adjusting the Bone Angle (BA) as follows:
\begin{align}
\hat{\boldsymbol{B}'} = \hat{\boldsymbol{B}} + \boldsymbol{\gamma_{ba}}, && \text{(BA operation).}
\end{align} 
Specifically, BA operation will rotate the input bone direction vector $\hat{\boldsymbol{B}}$ by $\boldsymbol{\gamma_{ba}}$, generating a new bone direction vector $\hat{\boldsymbol{B}'}$.

\item $\boldsymbol{\gamma_{bl}} \in \mathbb{R}^{1\times (J-1)}$ {represents the bone length ratio vector} that is used for adjusting the {Bone Length (BL)}:
\begin{align}
\|\boldsymbol{B}'\| = \|\boldsymbol{B}\| \times (1+\boldsymbol{\gamma_{bl}}), && \text{(BL operation).}
\end{align} 
BL operation modifies the input bone length vector $\|\boldsymbol{B}\|$ by $\boldsymbol{\gamma_{bl}}$ to adjust the body size. Notably, to ensure bio-mechanical symmetry, the left and right body parts share the same parameters.

\item  $\boldsymbol{R}\in \mathbb{R}^{3\times3}$ and $\boldsymbol{t}\in \mathbb{R}^{3\times1}$ denote the rotation and translation parameters respectively for Rigid Transformation (RT) operation to control pose view point and position:
\begin{align}
\boldsymbol{X}' = \boldsymbol{R} [\mathcal{H}^{-1}(\boldsymbol{B}')] + \boldsymbol{t}, && \text{(RT operation),}
\end{align} 
where $\boldsymbol{B}'=\|\boldsymbol{B}'\| \times \hat{\boldsymbol{B}'}$ is the augmented bone vector from the above BA and BL operations. $\mathcal{H}^{-1}$ is the inverse hierarchical conversion to transform $\boldsymbol{B}'$ back to a 3D pose~\cite{wandt2019repnet,Li_2020_CVPR}.
\end{enumerate}
By applying these operations, the augmentor can generate the augmented 3D pose $\boldsymbol{X}'$ with more challenging pose, body size, view point and position from the original 3D pose $\boldsymbol{X}$ (Fig.~\ref{fig:augmentor}).
The augmented pose is then re-projected to 2D with $\boldsymbol{x}'=\Pi(\boldsymbol{X}')$, where $\rm{\Pi}: \mathbb{R}^3 \to \mathbb{R}^2$ denotes perspective projection~\cite{Hartley2003MVG} via the camera parameters from the original data.
The augmented  2D-3D pair $\{\boldsymbol{x}', \boldsymbol{X}'\}$ is then used for further training the pose estimator. 

\begin{figure}[!t]
\centering
\includegraphics[width=0.85\linewidth]{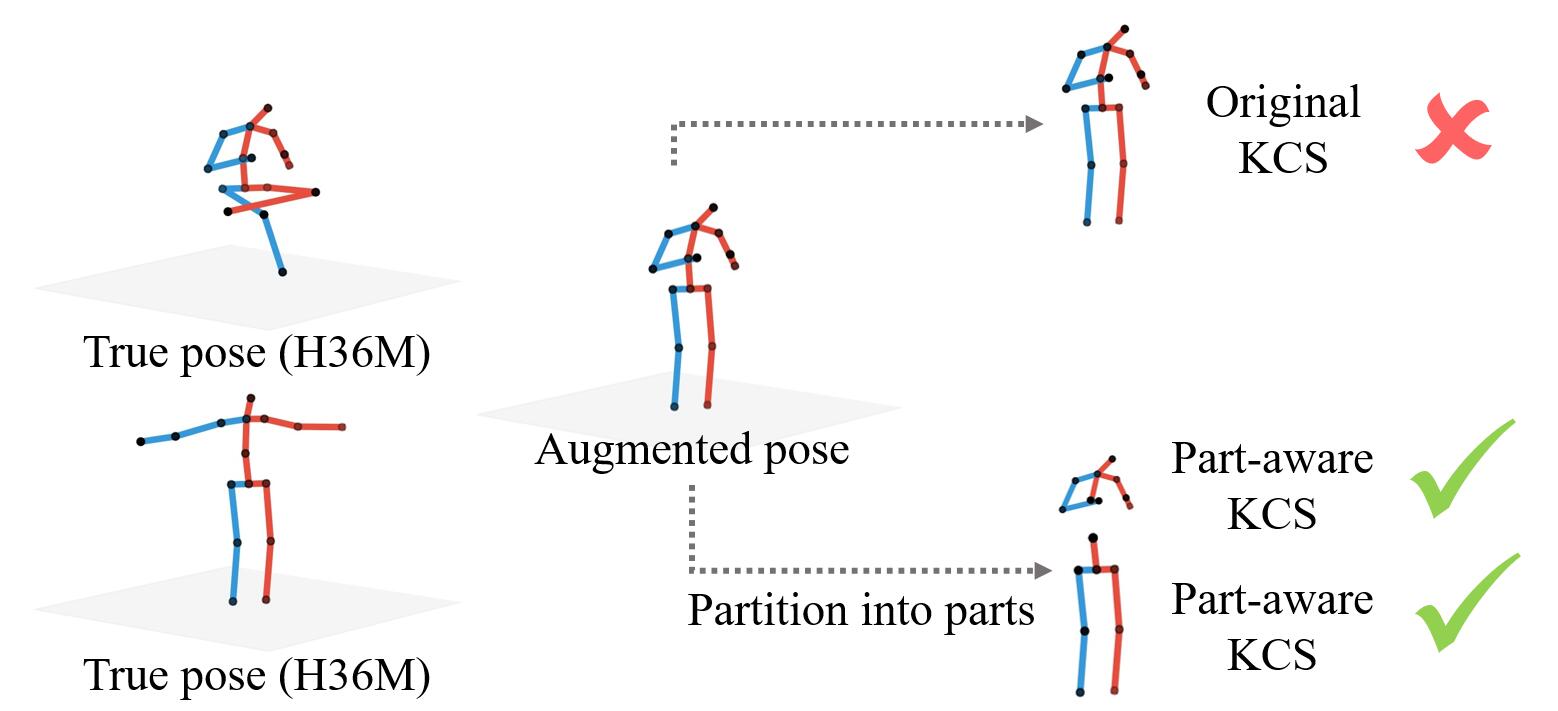}
\caption{\textbf{Illustrations of the difference between original and part-aware KCS based discriminator.} 
{Given a novel and valid augmented pose, the original KCS based discriminator would wrongly classify it as fake as it does not appear in source data (H36M), while the part-aware KCS based discriminator would recognize is as real and approve it, since it inspects local joint relations. 
It can be seen the part-aware KCS based discriminator can help the augmentor generate more diverse and plausible pose augmentation.}} 
\label{fig:d3d}
\end{figure}

\myparagraph{Discriminator}
Due to lacking priors in the augmentation procedure, the augmented poses may present implausible joint angles that violate the bio-mechanical structure~\cite{akhter2015joint_angle_limit}, or unreasonable positions and view points. 
Though such poses are indeed harder cases for the estimator, training on them would not benefit the model generalization ability.

To ensure the plausibility of the augmented poses, we introduce a pose discriminator module to guide the augmentation.
Specifically, the module consists of a 3D pose  discriminator $D_{3d}$ for evaluating the joint angle plausibility and a 2D discriminator $D_{2d}$ for evaluating the body size, viewpoint and position plausibility. 

The key to the 3D pose discriminator design is to ensure the pose plausibility without sacrificing the diversity.  
Inspired by the Kinematic Chain Space (KCS)~\cite{wandt2019repnet}, we design a \textit{part-aware} KCS as input to the discriminator. Instead of taking the whole body pose into consideration as in the original KCS, our part-aware KCS only focuses on local joint angle and thus enlarges the feasible region of the augmented pose, ensuring both plausibility and diversity (Fig.~\ref{fig:d3d}).

Specifically, to compute the part-aware KCS of an input pose, either $\boldsymbol{X}$ or its augmentation $\boldsymbol{X}'$,  we convert the pose to its bone direction vector $\hat{\boldsymbol{B}}$ as above and separate it into 5 parts (torso and left/right arm/leg)~\cite{akhter2015joint_angle_limit},  denoted as $\hat{\boldsymbol{B}}_{i}, i=1,\ldots,5,$ respectively.
We then calculate the following local joint angle matrix $KCS^i_{local}$ for the $i$-th part:
\begin{equation}
    KCS^i_{local} 
    = \hat{\boldsymbol{B}}_{i}^\top \hat{\boldsymbol{B}}_{i},
    \label{kcs}
\end{equation}
which encapsulates the inter joint angle information within the $i$-th part.
Based on the above local KCS representation, a 3D pose discriminator $D_{3d}$ is constructed which takes the $KCS^i_{local}$ as input and is trained for distinguishing the original and augmented 3D poses.  

Besides the 3D discriminator, we also introduce a 2D discriminator to guide the augmentor to generate real body size, view points and positions. 
As the 2D poses contain information such as view point (rotation), position (translation), and body size (bone length), the 2D discriminator can learn such information through adversarial training
and guide the pose augmentor in generating realistic rotation $\boldsymbol{R}$, translation $\boldsymbol{t}$, and bone length ratio $\boldsymbol{\gamma_{bl}}$.

\myparagraph{Estimator}
The pose estimator $\mathcal{P}$ estimates 3D poses from 2D poses. We use the original and augmented 2D-3D pose pair $\{\boldsymbol{x}, \boldsymbol{X}\}$ and $\{\boldsymbol{x}', \boldsymbol{X}'\}$ to train the pose estimator. 
The pose estimator contains a feature extractor to capture internal features from 2D poses, and a regression layer to estimate the corresponding 3D poses.
Moreover, any existing effective estimator can be implemented in our PoseAug framework. 
In Sec.~\ref{sec:different-estimators}, we conduct experiments to check robustness of PoseAug with different estimators, and the results show PoseAug can bring noticeable improvements on both source and cross-scenario datasets for all models.

\subsection{Training Loss} \label{sec:training-loss}

\myparagraph{Pose estimation loss}
We adopt the mean squared errors (MSE) of the ground truth (GT) $\boldsymbol{X}$ and predicted poses $\widetilde{\boldsymbol{X}}$ as the pose estimation loss, which is formulated as
\begin{equation}
\mathcal{L}_{\mathcal{P}} = \|\boldsymbol{X}-\widetilde{\boldsymbol{X}}\|_2^2 \label{eq:poseloss}.
\end{equation}
We train the pose estimator using $\mathcal{L}_{\mathcal{P}}$ with both original and augmented pose pairs jointly, which can significantly boost performance for the challenging in-the-wild scenes.

\myparagraph{Pose augmentation loss}
To facilitate model training, augmented data should harder than the original one, \ie, $\mathcal{L}_{\mathcal{P}}(\boldsymbol{X}') > \mathcal{L}_{\mathcal{P}}(\boldsymbol{X})$, but not too hard to hurt the training process. 
A simple way to design the loss function is to let the difference between the pose estimation loss on augmented and original data within a proper range. 
Inspired by~\cite{mao2017least,li2020pointaugment}, we implement a controllable feedback loss as
\begin{align}
\mathcal{L}_{fb} = |1.0 - \exp[\mathcal{L}_{\mathcal{P}}(\boldsymbol{X}') - \beta\mathcal{L}_{\mathcal{P}}(\boldsymbol{X})]|, \label{eq:fb-loss}
\end{align}
where $\beta>1$ controls the difficulty level for the generated poses,  making the value of $\mathcal{L}_{\mathcal{P}}(\boldsymbol{X}')$ stay within a certain range w.r.t.\  $\mathcal{L}_{\mathcal{P}}(\boldsymbol{X})$.
During training, as the pose estimator becomes increasingly more powerful, we accordingly increase $\beta$ value to generate more challenging augmentation data for training it.

Additionally, to prevent extremely hard cases from causing training collapse, we introduce a rectified  L2 loss for regularizing the augmentation parameters $\boldsymbol{\gamma_{ba}}$ and $\boldsymbol{\gamma_{bl}}$:
\begin{equation}
  \mathcal{L}_{reg}(\boldsymbol{\gamma})=\begin{cases}
    0, & \text{if $\bar{\boldsymbol{\gamma}} < threshold$},\\
    \|\boldsymbol{\gamma}\|^2, & \text{otherwise}, \label{eq:regulation-term}
  \end{cases}
\end{equation}
where $\boldsymbol{\gamma}$ denotes { $\boldsymbol{\gamma_{ba}}$ and $\boldsymbol{\gamma_{bl}}$, 
and $\bar{\boldsymbol{\gamma}}$ denotes the mean value over all of its elements.}
Combining Eqn.~(\ref{eq:fb-loss}) and Eqn.~(\ref{eq:regulation-term}), the overall augmentation loss $\mathcal{L}_{\mathcal{A}}$ is formulated as
\begin{equation}
  \mathcal{L}_{\mathcal{A}} = \mathcal{L}_{fb} + \mathcal{L}_{reg}. \label{eq:aug-loss}
\end{equation}

\myparagraph{Pose discrimination loss}
For the discrimination loss $\mathcal{L}_{\mathcal{D}}$, we adopt the LS-GAN loss~\cite{mao2017least} for both 3D and 2D spaces:
\begin{equation}
\begin{split}
    \mathcal{L}_{\mathcal{D}}  = \mathbb{E}[(D_{3d}(\boldsymbol{X})-1)^2] &+ \mathbb{E}[D_{3d}(\boldsymbol{X}')^2]  \\
     + \mathbb{E}[(D_{2d}(\boldsymbol{x})-1)^2] & + \mathbb{E}[D_{2d}(\boldsymbol{x}')^2],
\end{split} \label{eq:d-loss}
\end{equation}
where $\{\boldsymbol{x}, \boldsymbol{X}\}$ and $\{\boldsymbol{x}', \boldsymbol{X}'\}$ denote the original (real) and the augmented (fake) pose pairs, respectively.

\myparagraph{End-to-end training strategy}
With the differentiable design, the pose augmentor, discriminator and estimator can be jointly trained end-to-end. 
We update them alternatively by minimizing losses Eqn.~\eqref{eq:aug-loss}, Eqn.~\eqref{eq:d-loss} and Eqn.~\eqref{eq:poseloss}. 
In addition, we first pre-train the pose estimator $\mathcal{P}$ before training the whole framework end-to-end, which ensures stable training and produces better performance.
\section{Experiments}

We study four questions in experiments.
1) Is PoseAug able to improve performance of 3D pose estimator for both intra-dataset and cross-dataset scenarios? 
2) Is PoseAug effective at enhancing diversity of training data?
3) Is PoseAug consistently effective for different pose estimators and cases with limited training data?
4) How does each component of PoseAug take effect?
We experiment on H36M, 3DHP and 3DPW. 
Throughout the experiments, unless otherwise stated we adopt  single-frame version of VPose~\cite{pavllo2019videopose3d} as pose estimator.

\subsection{Datasets}

\myparagraph{Human3.6M (H36M)}~\cite{ionescu2014human3}
Following previous works~\cite{martinez2017simple,zhao2019semantic}, we train our model on subjects S1, 5, 6, 7, 8 of H36M and evaluate on subjects S9 and S11. We use two evaluation metrics:  Mean Per Joint Position Error (MPJPE)  in millimeters and  MPJPE over aligned predictions with  GT 3D poses by a rigid transformation (PA-MPJPE).

\myparagraph{MPI-INF-3DHP (3DHP)}~\cite{mehta2017vnect}
It is a large 3D pose dataset with 1.3 million frames, presenting more diverse motions than H36M. 
We use its test set to evaluate the model's generalization ability to unseen environments, using metrics of MPJPE, Percentage of Correct Keypoints (PCK) and Area Under the Curve (AUC).

\myparagraph{3DPW}~\cite{vonMarcard2018} 
It is an in-the-wild dataset with more complicated motions and scenes. 
To verify generalization of the proposed method to challenging in-the-wild scenarios, we use its test set for evaluation with PA-MPJPE as metric.

\myparagraph{\textbf{MPII}~\cite{andriluka20142d} and \textbf{LSP}~\cite{johnson2010clustered}}
They are in-the-wild datasets with only 2D body joint annotations and used for qualitatively evaluating model generalization for unseen poses.

\subsection{Results}

\myparagraph{Results on H36M}
We compare PoseAug with state-of-the-art methods~\cite{zhao2019semantic,sharma2019monocular,pavllo2019videopose3d, moon2019camera,Li_2020_CVPR} on H36M. Similar to~\cite{Li_2020_CVPR}, we use 2D poses from HR-Net \cite{sun2019deep} as inputs. 
As shown in Table~\ref{tab:h36m_fully}, our method outperforms SOTA methods~\cite{zhao2019semantic,sharma2019monocular,pavllo2019videopose3d, moon2019camera} by a large margin, indicating its effectiveness. 
Notably, compared with the previous best augmentation method~\cite{Li_2020_CVPR}, our PoseAug achieves lower  MPJPE even though it uses external bone length data for data augmentation and nearly $3\times$ more data than ours for model training. This clearly verifies advantages of PoseAug's online augmentation scheme\textemdash it can generate more diverse  and informative data that better benefit model training. 

\begin{table}[h]
	\small
	\centering
	\setlength{\tabcolsep}{1mm}
	\vspace{-3mm}
	\begin{tabular}{l|cc}
		\specialrule{1pt}{1pt}{1pt}
		Method &  MPJPE~($\downarrow$) & PA-MPJPE~($\downarrow$)  \\
		\hline
        \rowcolor{grayLight}
		SemGCN (CVPR'19)~\cite{zhao2019semantic}  & 57.6 & -  \\		
		\rowcolor{grayDark}
		Sharma~\et (CVPR'19)~\cite{sharma2019monocular}   & 58.0 & 40.9 \\
		\rowcolor{grayLight}
		VPose (CVPR'19)~\cite{pavllo2019videopose3d} (1-frame)   & 52.7 & 40.9 \\
		\rowcolor{grayDark}
		Moon~\et (ICCV'19)~\cite{moon2019camera}  & 54.4 &-  \\
		\rowcolor{grayLight}
		Li~\et (CVPR'20)~\cite{Li_2020_CVPR} & 50.9 & \textbf{38.0} \\
		\hline
		\rowcolor{grayDark}
		Ours   & \textbf{50.2}  &39.1  \\	
		\specialrule{1pt}{1pt}{2pt}	
	\end{tabular}
	
	\caption{\textbf{Results on H36M} in terms of MPJPE and PA-MPJPE. Best results are shown in \textbf{bold}.}
	\label{tab:h36m_fully}

\end{table}

\begin{table}[h]
	\small
	\centering
	\setlength{\tabcolsep}{1.6mm}
	\vspace{-3mm}
	\begin{tabular}{l|c|ccc}
		\specialrule{1pt}{1pt}{2pt}
		Method & CE & PCK~($\uparrow$) & AUC~($\uparrow$) & MPJPE~($\downarrow$) \\
		\hline
		\rowcolor{grayLight}
		Mehta~\et~\cite{mono3dhp2017} & & 76.5 & 40.8 & 117.6 \\
		\rowcolor{grayDark}
		VNect~\cite{mehta2017vnect}  &  & 76.6 & 40.4 & 124.7 \\
		\rowcolor{grayLight}
		Multi Person~\cite{singleshotmultiperson2018} & & 75.2 & 37.8 & 122.2 \\
		\rowcolor{grayDark}
		OriNet~\cite{luo2018orinet} & & 81.8 & 45.2 & {89.4} \\
		\hline
		\rowcolor{grayLight}		
		LCN~\cite{ci2019optimizing} & \checkmark & 74.0 & 36.7 & -\\
		\rowcolor{grayDark}
		HMR~\cite{hmrKanazawa17} & \checkmark & 77.1 &	40.7 & 113.2 \\
		\rowcolor{grayLight}
		SRNet~\cite{zeng2020srnet} & \checkmark & 77.6 & 43.8 & -\\
		\rowcolor{grayDark}
		Li~\et~\cite{Li_2020_CVPR} & \checkmark & 81.2 &	46.1 & 99.7 \\
		\rowcolor{grayLight}
		RepNet~\cite{wandt2019repnet} & \checkmark & 81.8 & 54.8 & 92.5 \\
		\hline
		\rowcolor{grayDark}			
		Ours  & \checkmark & \textbf{88.6} & \textbf{57.3} & \textbf{73.0}\\
		\rowcolor{grayLight}			
		Ours(+Extra2D)  & \checkmark &  {89.2} & {57.9} & {71.1}\\
		\specialrule{1pt}{1pt}{2pt}
	\end{tabular}
	\caption{\textbf{Results on 3DHP}. CE denotes cross-scenario evaluation. PCK, AUC and MPJPE are used for evaluation. }
	\label{tab:3dhp}
	\vspace{-3mm}
\end{table}

%#####################################################################
\begin{figure*}[!h]
    \centering
    \includegraphics[width=0.95\linewidth]{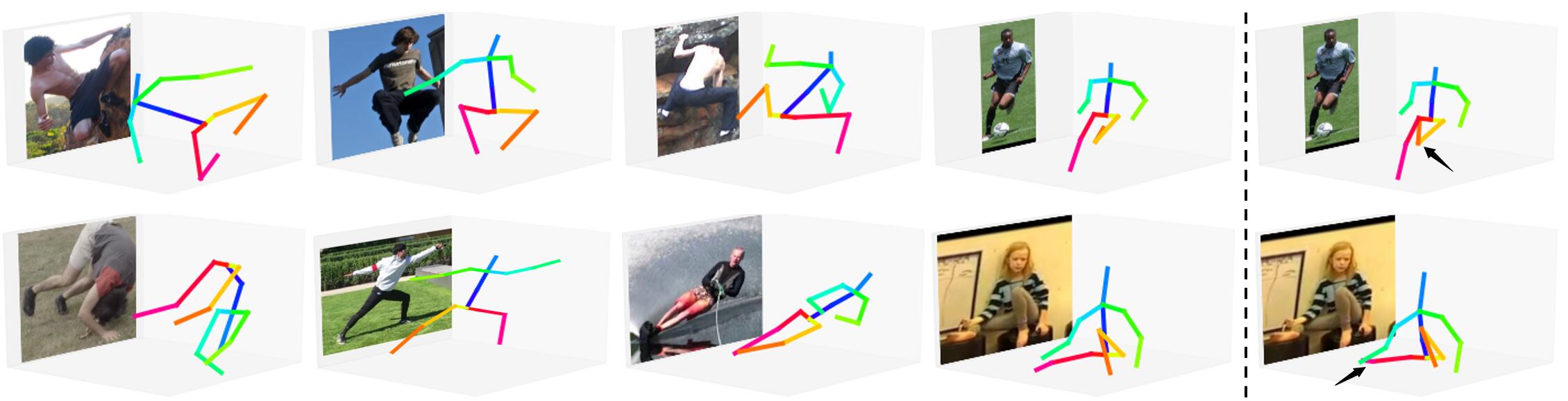}
    \caption{Example 3D pose estimations from LSP, MPII, 3DHP and 3DPW.
    Our results are shown in the left four columns. The rightmost column shows results of \textit{Baseline}\textemdash VPose~\cite{pavllo2019videopose3d} trained w/o PoseAug. Errors are highlighted by  black arrows.} 
    \label{fig:visualization}
    \vspace{-2mm}
\end{figure*}
%#####################################################################

\myparagraph{Results on 3DHP (cross-scenario)}
We then evaluate how PoseAug facilitates model generalization to cross-scenario datasets. 
We compare PoseAug against various state-of-the-art methods, including the latest one using offline data augmentation~\cite{Li_2020_CVPR}, the ones exploiting complex network architecture~\cite{ci2019optimizing,zeng2020srnet} and  weakly-supervised learning~\cite{hmrKanazawa17,wandt2019repnet} and the ones trained on the training set of 3DHP~\cite{mono3dhp2017,mehta2017vnect,singleshotmultiperson2018,luo2018orinet}. 
From Table~\ref{tab:3dhp},  we can observe our method achieves the best performance w.r.t.\ all the metrics, outperforming previous approaches by a large margin. 
This verifies the effectiveness of PoseAug in improving model generalization to unseen scenarios. 
Moreover, PoseAug can further improve the performance (from 73.0 to 71.1 in MPJPE) by using additional in-the-wild 2D poses (MPII) to train the 2D discriminator. 
This demonstrates its {extensibility} in leveraging  extra 2D poses to further enrich the diversity of augmented data.

\vspace{2mm}
\myparagraph{Results on 3DPW (cross-scenario)} 
We train four 3D pose estimators~\cite{martinez2017simple,zhao2019semantic,cai2019exploiting,pavllo2019videopose3d} without and with PoseAug on H36M and compare their generalization performance on 3DPW. 
As shown in Table~\ref{tab:ablation-estimator}, on average, PoseAug brings $12.6\%$ improvements for all the models.

\vspace{-2mm}
\begin{table}[!h]
\small
\centering
\newcommand{\TableEntry}[2]{\textbf{#1}~\scriptsize{\green{(-#2)}}}
\setlength{\tabcolsep}{4mm}
\caption{\small Results in PA-MPJPE for four estimators on 3DPW.}
\label{tab:ablation-estimator}
\vspace{-4mm}
\begin{tabular}{l|c}
\specialrule{1pt}{1pt}{1pt}
Method & PA-MPJPE~($\downarrow$) \\
\hline
\rowcolor{grayLight}
SemGCN~\cite{zhao2019semantic} & 102.0  \\
\rowcolor{grayDark}
+ PoseAug & \TableEntry{82.2}{19.8} \\
\hline
\rowcolor{grayLight}
SimpleBaseline~\cite{martinez2017simple} &  89.4  \\
\rowcolor{grayDark}
+ PoseAug & \TableEntry{78.1}{11.3}  \\
\hline
\rowcolor{grayLight}
ST-GCN~\cite{cai2019exploiting}(1-frame) & 98.0  \\
\rowcolor{grayDark}
+ PoseAug & \TableEntry{73.2}{24.8}  \\
\hline
\rowcolor{grayLight}
VPose~\cite{pavllo2019videopose3d} (1-frame) & 94.6  \\
\rowcolor{grayDark}
+ PoseAug & \TableEntry{81.6}{13.0} \\
\specialrule{1pt}{1pt}{1pt}
\end{tabular}
\end{table}

\vspace{-2mm}
\myparagraph{Qualitative results}
For subjective evaluation, we choose four challenging datasets, \ie, LSP, MPII, 3DHP and 3DPW, with large varieties of postures, body sizes, and view points between their data and the data from H36M. 
Results are shown in Fig.~\ref{fig:visualization}. 
We can see our method performs fairly well, even for those unseen difficult poses.

\vspace{-2mm}
\begin{table*}[!t]
\small
\centering
\newcommand{\TableEntry}[2]{\textbf{#1}~\scriptsize{\green{(-#2)}}}
\setlength{\tabcolsep}{2mm}
\vspace{-3mm}
\begin{tabular}{l|cccc|cccc}
\specialrule{1pt}{1pt}{1pt}
\multicolumn{1}{c|}{} & \multicolumn{4}{c|}{H36M}  & \multicolumn{4}{c}{3DHP} \\ 
\hline
Method & \multicolumn{1}{c}{DET} & \multicolumn{1}{c}{CPN} & \multicolumn{1}{c}{HR} & \multicolumn{1}{c|}{GT} &  \multicolumn{1}{c}{DET} & \multicolumn{1}{c}{CPN} & \multicolumn{1}{c}{HR} & \multicolumn{1}{c}{GT}\\ 
\hline
\rowcolor{grayLight}
SemGCN~\cite{zhao2019semantic} & 67.5 & 64.7 & 57.5 & 44.4 & 101.9 & 98.7 & 95.6 & 97.4 \\
\rowcolor{grayDark}
+ PoseAug & \TableEntry{65.2}{2.3}  & \TableEntry{60.0}{4.8} & \TableEntry{55.0}{2.5} & \TableEntry{41.5}{2.8} & \TableEntry{89.9}{11.9} & \TableEntry{89.3}{9.4}  & \TableEntry{89.1}{6.5}  & \TableEntry{86.1}{11.2} \\
\hline
\rowcolor{grayLight}
SimpleBaseline~\cite{martinez2017simple} & 60.5 & 55.6 & 53.0 & 43.3 & 91.1 & 88.8 & 86.4 & 85.3 \\
\rowcolor{grayDark}
+ PoseAug & \TableEntry{58.0}{2.5} & \TableEntry{53.4}{2.2} & \TableEntry{51.3}{1.7} & \TableEntry{39.4}{3.9} & \TableEntry{78.7}{12.4} & \TableEntry{78.7}{10.1} & \TableEntry{76.4}{10.1} & \TableEntry{76.2}{9.1} \\
\hline
\rowcolor{grayLight}
ST-GCN~\cite{cai2019exploiting} (1-frame) &61.3 & 56.9 & 52.2 & 41.7 & 95.5 & 91.3 & 87.9 & 87.8 \\
\rowcolor{grayDark}
+ PoseAug &\TableEntry{59.8}{1.5} & \TableEntry{54.5}{2.4} & \TableEntry{50.8}{1.5} & \TableEntry{36.9}{4.8} & \TableEntry{83.5}{12.1} & \TableEntry{77.7}{13.6} & \TableEntry{76.6}{11.3} & \TableEntry{74.9}{12.9} \\
\hline
\rowcolor{grayLight}
VPose~\cite{pavllo2019videopose3d} (1-frame) & 60.0 & 55.2 & 52.7 & 41.8 & 92.6  & 89.8 & 85.6 & 86.6 \\
\rowcolor{grayDark}
+ PoseAug & \TableEntry{57.8}{2.2} & \TableEntry{52.9}{2.3} & \TableEntry{50.2}{2.5} & \TableEntry{38.2}{3.6} & \TableEntry{78.3}{14.4} & \TableEntry{78.4}{11.4} & \TableEntry{73.2}{12.4} & \TableEntry{73.0}{13.6}\\
\specialrule{1pt}{1pt}{1pt}
\end{tabular}
\caption{{Performance comparison in MPJPE for various pose estimators trained w/o and with PoseAug on H36M and 3DHP datasets. DET, CPN, HR and GT denote 3D pose estimation model trained on different 2D pose sources, respectively.
We evaluate the model on H36M test set with the corresponding 2D pose sources. 
On 3DHP test set, we use GT 2D poses as input for evaluating model's generalization. We can observe PoseAug consistently decreases errors for all datasets and estimators.} 
}
\label{tab:ablation-estimator}
\end{table*}

\subsection{Analysis on PoseAug}
\myparagraph{Applicability to different estimators}
\label{sec:different-estimators}
Our PoseAug framework is generic and applicable to different 3D pose estimators.
To demonstrate this, we employ four representative 3D pose estimators as backbones: 1) SemGCN~\cite{zhao2019semantic}, a graph-based 3D pose estimation network; 2) SimpleBaseline~\cite{martinez2017simple}, an effective MLP-based network; 3) ST-GCN~\cite{cai2019exploiting} (1-frame), a pioneer network that uses GCN-based architecture to encode global and local joint relations; and 4) VPose~\cite{pavllo2019videopose3d} (1-frame), a fully-convolutional network with SOTA performance. 
We train these models on the H36M dataset using 2D poses from four different 2D pose detectors, including CPN~\cite{chen2018cpn}, DET~\cite{Detectron2018}, HR-Net~\cite{sun2019deep} and groundtruth (GT).
We evaluate these models on the test set of H36M and 3DHP w.r.t. MPJPE metric. 
On H36M, we use the corresponding 2D poses for evaluation; while on 3DHP, we evaluate these models with GT 2D poses to filter out the influence of 2D pose detectors.
The results are shown in Table~\ref{tab:ablation-estimator}. 
We can see PoseAug brings clear improvements to all models on both H36M and more challenging 3DHP datasets. 
Notably, they obtain more than 13.1\% average improvement on 3DHP when trained with PoseAug.  
\vspace{-4mm}
\begin{figure}[!h]
  \centering
  \subfloat[{\plottitle ~~~~Source dataset: H36M}]{\includegraphics[width=0.5\linewidth]{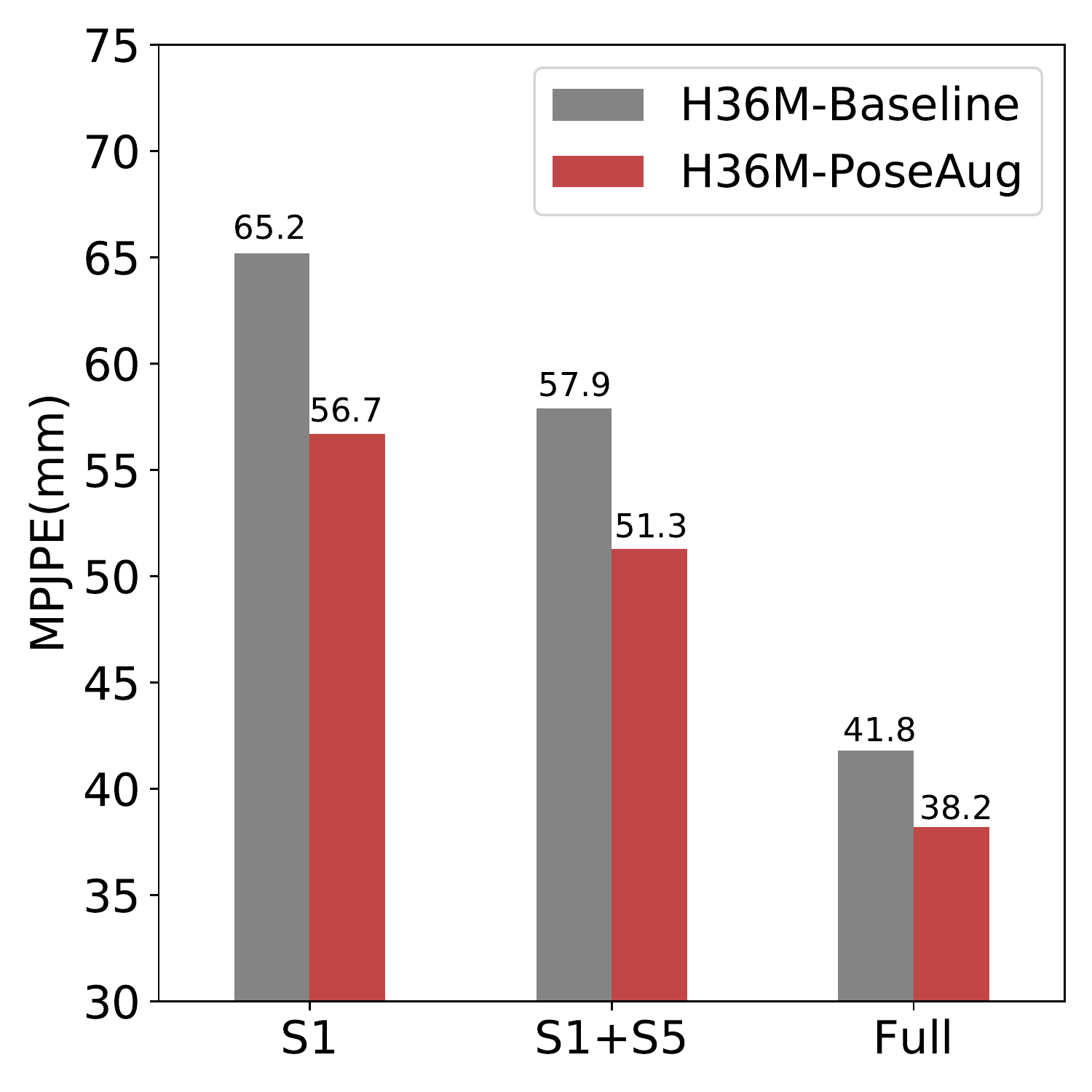}}
  \subfloat[{\plottitle ~~~~Cross dataset: 3DHP}]{\includegraphics[width=0.5\linewidth]{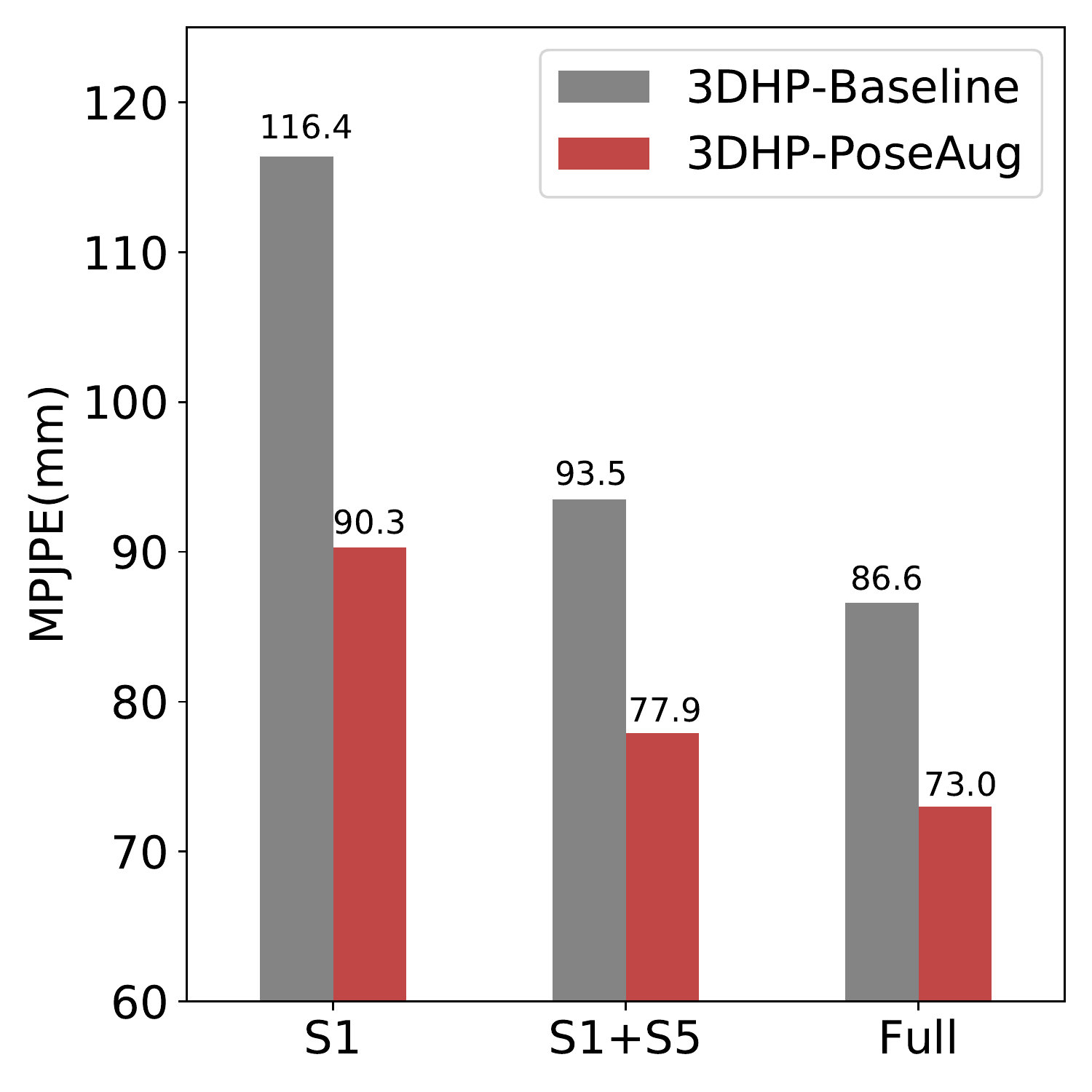}}
  \caption{Ablation study on limited data setup. 
  We report MPJPE for evaluation. Best viewed in color.
  }
  \vspace{-2mm}
  
\label{fig:limite_data}
\end{figure}

\myparagraph{Effectiveness for limited training data cases}
3D pose annotations are expensive to collect, making limited training data a common challenge.
To demonstrate the effectiveness of our method on addressing such cases,  we use pose data from H36M S1 and S1+S5 for model training which only contain 16\% and 41\% training samples, respectively. 
The results in Fig.~\ref{fig:limite_data} show PoseAug consistently improves model performance with varying amounts of training data, on both H36M and 3DHP. 
Meanwhile, the improvements brought by our method are more significant for cases with less training data (\eg, MPJPE in 3DHP, S1: 116.4 $\rightarrow$ 90.3, Full: 86.6 $\rightarrow$ 73.0). 
Moreover, in cross-scenario generalization, our method trained with only S1 achieves the comparable result (MPJPE: 90.3) to baseline trained using full dataset (MPJPE: 86.6), and our method trained with S1+S5 can outperform baseline trained using full dataset by a large margin (77.9 vs 86.6 in MPJPE). 

\myparagraph{Analysis on the augmentor}
We then check the effectiveness of each module in augmentor.
Table~\ref{tab:ablation-augmentor} summarizes the results. 
By gradually adding the BA, RT and BL operations, the pose estimation error can  be monotonically decreased from 41.8/86.6 to 38.8/73.5 (on H36M/3DHP).
Moreover, incorporating the error feedback guidance can further improve performance to 38.2 for H36M and 73.0 for 3DHP. 
These verify the effectiveness of each module of the augmentor in producing more effective augmented samples. 
Among these modules, RT contributes the most to cross-scenario performance, which implies it benefits data diversity most effectively.

\vspace{-2mm}
\begin{table}[h]
	\small
	\centering
	\setlength{\tabcolsep}{1.3mm}
    \newcommand{\TableEntry}[2]{{#1}~\scriptsize{\green{(-#2)}}}
	\vspace{-3mm}
	\begin{tabular}{l|cccc|cc}
		\specialrule{1pt}{1pt}{2pt}
		Method &  BA & RT & BL & Feedback  &  H36M~($\downarrow$) & 3DHP~($\downarrow$)\\
		\hline 
		\rowcolor{grayLight}
		Baseline &  &  &  &  &   41.8 & 86.6  \\
		\rowcolor{grayDark}
		Variant A & \checkmark &  &  &  &   \TableEntry{39.7}{2.1} & \TableEntry{85.2}{1.4}  \\
		\rowcolor{grayLight}
		Variant B &  & \checkmark &  &  &   \TableEntry{39.2}{2.6} & \TableEntry{75.9}{10.7}  \\
		\rowcolor{grayDark}
		Variant C & \checkmark & \checkmark &  &    & \TableEntry{39.1}{2.7} & \TableEntry{75.5}{11.1}  \\
		\rowcolor{grayLight}
		Variant D & \checkmark & \checkmark & \checkmark   &  & \TableEntry{38.8}{3.0} & \TableEntry{73.5}{13.1}  \\
		\rowcolor{grayDark}
		PoseAug & \checkmark & \checkmark & \checkmark & \checkmark&   \TableEntry{\textbf{38.2}}{3.6} & \TableEntry{\textbf{73.0}}{13.6}  \\
		\specialrule{1pt}{1pt}{2pt}	
	\end{tabular}
	\caption{Ablation study on components of the augmentor. We report MPJPE on H36M and 3DHP datasets.}
% 	\vspace{-4mm}
	\label{tab:ablation-augmentor}
\end{table}

%#################################
\begin{figure}[!t]
\centering
\includegraphics[width=1\linewidth]{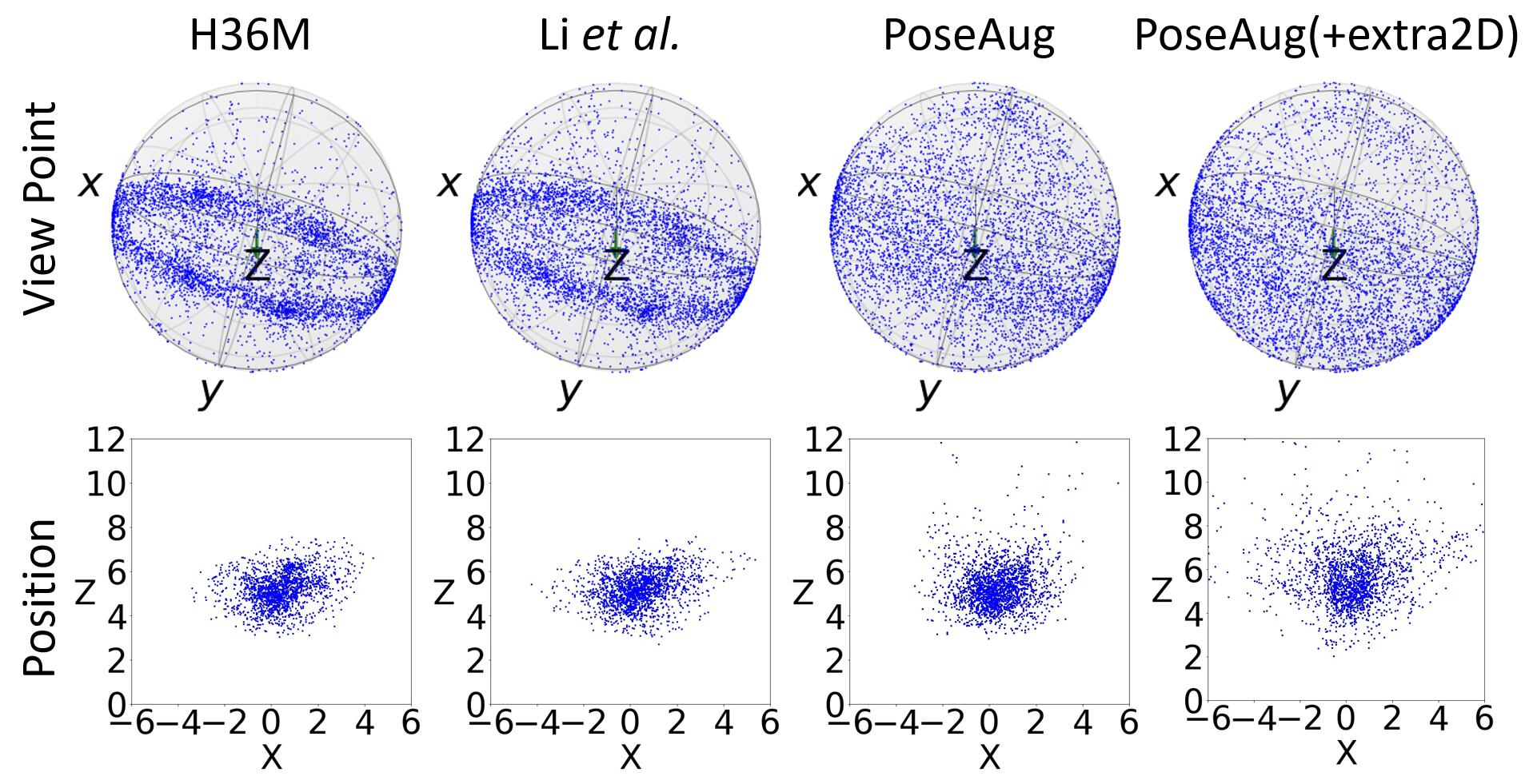}
\caption{Distribution on view point (top row) and position (bottom row) for original data H36M, and augmented data from Li~\et~\cite{Li_2020_CVPR}, PoseAug (3rd column) and PoseAug with extra 2D poses.
This distribution shows PoseAug significantly improves diversity of view point and position.}
\vspace{-2mm}
\label{fig:rt_diversity}
\end{figure}
%#################################

% \vspace{-2mm}
\myparagraph{Analysis on diversity improvement}
{To demonstrate effectiveness of PoseAug in enhancing data diversity, considering RT operation which augments the view point and position contributes the most to cross-scenario performance, as shown in Table~\ref{tab:ablation-augmentor}, we make diversity analysis on view point and position distribution. Fig.~\ref{fig:rt_diversity} demonstrates the distributions of view point and position of H36M and the augmented data generated by Li~\et~\cite{Li_2020_CVPR} and our method.}
For H36M data, one can observe their view points concentrate near to the xz-plane with a limited diversity along the y-axis; and their positions form a small and concentrated cluster, also showing a limited diversity. 
This explains why the model trained on H36M hardly generalizes to in-the-wild scenarios. 
Similarly, we observe small divergence 
for the view point and position distribution of augmented data from Li~\et~\cite{Li_2020_CVPR}. This implies the diversity improvement from the handcrafted rule is limited. 
Comparably, our PoseAug can offer more plausible view points and positions using the learnable augmentor, with a much greater diversity.
In addition, the diversity on human positions can be further improved with extra 2D poses, 
which also explains its resulted improved generalization ability in Table~\ref{tab:3dhp}.

\vspace{2mm}
\myparagraph{Analysis on the discriminator} 
{We here demonstrate the effectiveness of plausibility guidance from the 2D and 3D discriminators. 
Table~\ref{tab:ablation-d23d} summarizes the results. 
By adding one of the 2D or 3D discriminators, the performance of baseline can be boosted by  {2.2/5.8 and 2.2/7.0} on H36M/3DHP, respectively.
Including both discriminators into PoseAug training can further boost the performance {by 3.6/13.6 on H36M/3DHP}, which clearly verify the effectiveness of both discriminators and also the importance of plausibility (in augmented poses) for estimator performance. 
}

% \vspace{-3mm}
\begin{table}[h]
	\small
	\centering
	\setlength{\tabcolsep}{2mm}
	\newcommand{\TableEntry}[2]{{#1}~\scriptsize{\green{(-#2)}}}
	\vspace{-3mm}
	\begin{tabular}{l|cc|cc}
		\specialrule{1pt}{1pt}{2pt}
		Method & ${\mathcal{D}_{2D}}$ & ${\mathcal{D}_{3D}}$ & H36M~($\downarrow$) & 3DHP~($\downarrow$)\\
		\hline 
		\rowcolor{grayLight}
		Baseline & & & 41.8 & 86.6 \\
		\rowcolor{grayDark}
		Variant A & \checkmark & &  \TableEntry{39.6}{2.2} & \TableEntry{80.8}{5.8}  \\
		\rowcolor{grayLight}
		Variant B & & \checkmark &  \TableEntry{39.6}{2.2} & \TableEntry{79.6}{7.0}  \\
		\rowcolor{grayDark}
		PoseAug & \checkmark & \checkmark &  \TableEntry{\textbf{38.2}}{3.6} & \TableEntry{\textbf{73.0}}{13.6}  \\
		\specialrule{1pt}{1pt}{2pt}	
	\end{tabular}
	\caption{
	Ablation study on the discriminators $\mathcal{D}_{2D}$ and $\mathcal{D}_{3D}$ on H36M and 3DHP. MPJPE is used for evaluation.
	}
	\label{tab:ablation-d23d}
\end{table} 

% \vspace{-2mm}
\myparagraph{Analysis on  part-aware KCS (PA-KCS)} 
To verify its effectiveness, we replace it in PoseAug with KCS~\cite{wandt2019repnet}.
Table~\ref{tab:ablation-discriminator} summarizes the results. PA-KCS clearly outperforms KCS on both 3DHP and 3DPW. This verifies our PA-KCS provides better guidance than KCS during training.

% \vspace{-2mm}
\begin{table}[h]
	\small
	\centering
	\setlength{\tabcolsep}{1.3mm}
    \newcommand{\TableEntry}[2]{{#1}~\scriptsize{\green{(-#2)}}}
    \caption{\small Ablation study on part-aware KCS (PA-KCS). We report MPJPE on 3DHP and PA-MPJPE on 3DPW.}
	\label{tab:ablation-discriminator}
	\vspace{-3mm}
	\begin{tabular}{l|cc|cc}
		\specialrule{1pt}{1pt}{2pt}
		Method & KCS & PA-KCS  &
		3DHP~($\downarrow$) &  3DPW~($\downarrow$)\\
		\hline 
		\rowcolor{grayLight}
		Baseline & & & 86.6 & 94.6 \\
		\rowcolor{grayDark}
		Variant A & \checkmark & & 
		\TableEntry{77.7}{8.9} & \TableEntry{88.4}{6.2} \\
		\rowcolor{grayLight}
		PoseAug & & \checkmark & 
		\TableEntry{\textbf{73.0}}{13.6} & \TableEntry{\textbf{81.6}}{13.0} \\
		\specialrule{1pt}{1pt}{2pt}	
	\end{tabular}
	\vspace{-4mm}
\end{table}

%%%%%%%%%% conclusion
\section{Conclusion}

In this paper, we develop an auto-augmentation framework, PoseAug, that learns to enrich the diversity of training data and improves performance of the trained pose estimation models.
The PoseAug effectively integrates three components including the augmentor, estimator and discriminator and makes them fully interacted with each other.
Specifically, the augmentor is designed to be differentiable and thus can learn to change 
major geometry factors of the 2D-3D pose pair to suit the estimator better by taking its training error as feedback.
The discriminator can ensure the plausibility of augmented data based on a novel part-aware KCS representation.
Extensive experiments justify PoseAug can augment diverse and informative data to boost estimation performance for various 3D pose estimators.

% \vspace{-2mm}
\myparagraph{Acknowledgement}
This research was partially supported by AISG-100E-2019-035, MOE2017-T2-2-151, NUS\_ECRA\_FY17\_P08 and CRP20-2017-0006. JZ would like to acknowledge the support of NVIDIA AI Tech Center (NVAITC) to this research project.

{\small
\bibliographystyle{ieee_fullname}
\bibliography{egbib}
}

\end{document}